# An Ensemble of Neural Networks for Non-Linear Segmentation of Overlapped Cursive Script

Amjad Rehman[1]

[1]College of Computer and Information Systems Al Yamamah University Riyadh 11512 Saudi Arabia

**Abstract**

Precise character segmentation is the only solution towards higher Optical Character Recognition (OCR) accuracy. In cursive script, overlapped characters are serious issue in the process of character segmentations as characters are deprived from their discriminative parts using conventional linear segmentation strategy. Hence, non-linear segmentation is an utmost need to avoid loss of characters parts and to enhance character/script recognition accuracy. This paper presents an improved approach for non-linear segmentation of the overlapped characters in handwritten roman script. The proposed technique is composed of a sequence of heuristic rules based on geometrical features of characters to locate possible non-linear character boundaries in a cursive script word. However, to enhance efficiency, heuristic approach is integrated with trained ensemble neural network validation strategy for verification of character boundaries. Accordingly, correct boundaries are retained and incorrect are removed based on ensemble neural networks vote. Finally, based on verified valid segmentation points, characters are segmented non-linearly. For fair comparison CEDAR benchmark database is experimented. The experimental results are much better than conventional linear character segmentation techniques reported in the state of art. Ensemble neural network play vital role to enhance character segmentation accuracy as compared to individual neural networks.

## 1. Introduction and background

The segmentation and recognition of cursive handwritten words has been an area of great interest from last few decades [1-10]. Despite of successes achieved in individual character recognition, the recognition of offline unconstrained cursive script is still fresh and yields poor accuracy results [11-15]. It is mainly due to the bottleneck problem inaccurate character segmentation of cursive script words into separated characters. Segmentation problem persists nearly as long as recognition problem. Higher segmentation accuracy causes good recognition rates [16-20]. Therefore, the segmentation is an essential component in any practical handwritten recognition system. The high accuracy in character segmentation is still of worth importance in script classification process [21-27]. Hence, enhanced and fast character segmentation techniques are still in demand. To avoid segmentation problem word level/holistic approaches are suggested [28-33]. While these approaches do avoid the difficulty of character segmentation issue, however, are suitable for the limited lexicon environments such as postal code searching, postal address, bank checks amount and so forth.

In the literature, most of character segmentation strategies concentrate on human writing to derive common rules [34-40]. Sometimes the derived rules are satisfactory, but most of the time they do not produce accurate results. Since human writing is not constant art rather it depends on various factors such as speed of writing, mood of the writer, sex difference and environment. On the other hand, researchers have employed various intelligent techniques to enhance segmentation accuracy of cursive handwritten words [41-46]. Above all, following identification of character boundaries, all approaches reported in the state of art perform linear/vertical segmentation. It has two major drawbacks, first characters lose their discriminator parts, second; it reduces classification accuracy due to the character's deprived parts [47-50].

In this paper, an enhanced cursive script non-linear segmentation scheme based on set of heuristic rules is proposed in order to determine letter boundaries and to avoid their discriminative parts removal. However, due to inherited features of cursive script over-segmentation is caused for few characters. To avoid over segmentation an ensemble neural model is integrated with this approach. Heuristically identified segmentation points are fed to the trained ANN for identification of correct/incorrect character boundaries. Prior to this, ANN is ensemble neural model with significant number of correct and incorrect letter boundaries. Finally, non-linear character segmentation is performed to segment characters for their further classification.

The rest of paper is composed of three main sections. Section 2, presents the segmentation methodology, section 3 presents training and testing of neural network to identify invalid character segmentation emerged from heuristics segmentation and based on valid segmentation points non-linear segmentation is performed. Last section 4 concludes the research.

## 2. Proposed Non-linear Segmentation Approach
   a. **Preprocessing**

Pre-processing is mandatory step in all image processing applications. It remove unnecessary detail and smooth line further processing, additionally, it speed up processing as few signals are to be processed. In this research noise removal and thinning operations are performed on scrip images. Initially, the input gray image is first converted to binary image by employing Otsu algorithm; slant corrected and is converted to skeleton format [51-53]. Accordingly, the selected images are preprocessed as exhibited in Figure 1(a).

   b. **Over-segmentation**

The process of over-segmentation is acquired to mark all possible character boundaries. However, unwanted/incorrect segmentations are also emerged as byproduct. In this research, the preprocessed images are over-segmented heuristically at distance d to mark all possible valid character boundaries as exhibited in Figure 1(b). Where, $d = w/n$ & w is the width of an image and n is determined heuristically.

   c. **Loop determination**

Loops/semi loops are always part of characters. Hence, to avoid loop segmentation, an enhanced criterion is proposed, accordingly, for each vertical line, crossing of foreground pixels are counted. If count is more than 1 then it is assumed it is crossing loop/semi loop therefore, vertical segmentation is termed as invalid and removed from images immediately as shown in Figure 1(c).

   d. **Character width**

Character width is another heuristic criterion to estimate character boundaries. Accordingly, if consecutive vertical lines are at standard width 'w', first and last vertical segmentation lines are included and in between are excluded as presented in Figure1 (d).

**e. Character boundaries**

Character width w is also adopted as true segmentation criteria. Accordingly, if vertical segment lines are at distance less than character standard width w, set their center as letter boundaries, as exhibited in Figure 1(e).

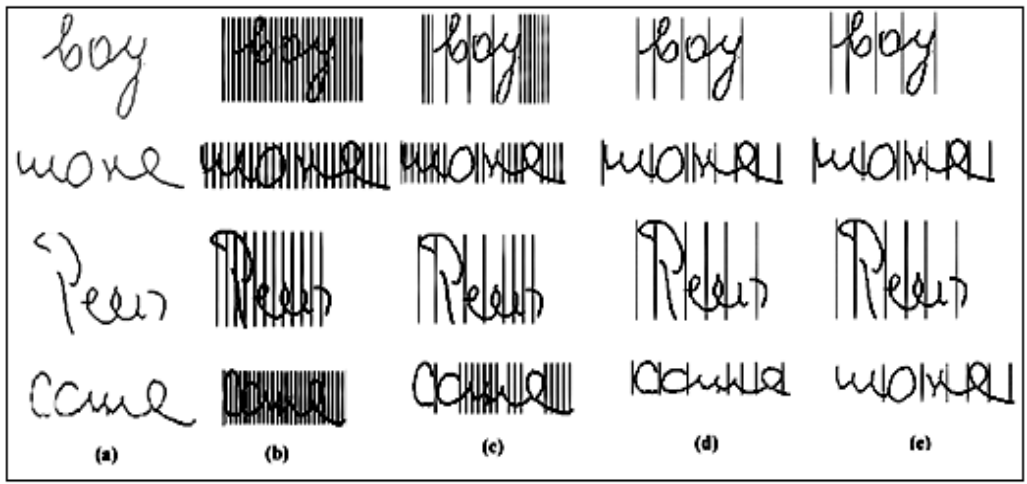

**Fig.1** Step wise results illustration of heuristic approach

Heuristic segmentation approach performed well in most cases as shown in figure 2(a). However, in case of few characters such as 'w', 'u', 'm', 'n' and 'v', over-segmentation occurred and heuristic rules are failed to find correct character boundaries in cursive script figure 2(b). Hence, to enhance segmentation accuracy, the results of heuristic segmentation are fed to a trained individual and ensemble neural networks in order to identify invalid segmentation detail in next section. Nonetheless, over-segmentation is least and emerged for few characters only, hence it caused classifier less burdened and fast.

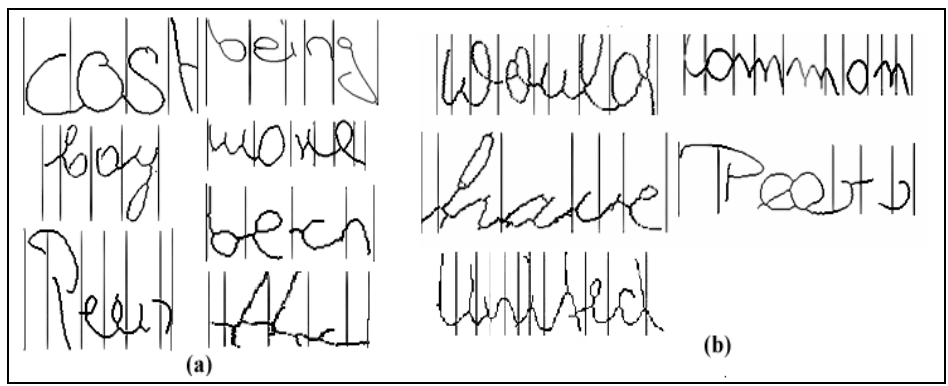

Fig. 2. Character segmentation using heuristics only (a) Successful (b) Failure results

**2.1 Ensemble of Neural Networks**

Neural networks are particularly suitable for the non-linear problems where traditional techniques are failed [54,55]. However, using neural networks a relationship could be drawn between inputs and outputs; hence these features are particularly suitable to assist in the character segmentation process. Literature is evident that ensemble of neural networks improve accuracy as compared to the individual network [56-60]. Accordingly, logic behind the ensemble of neural networks is to train few neural networks separately and finally their decisions are averaged to have a better decision. The output of ensemble network is a mean value of the output of all individual networks and output of ensemble network is normally better than individuals [61,63].

Saba et al., [64-66] emphasize that classification using single neural model results into low accuracy as compared to ensemble models. Accordingly, current research ensemble MLP and RBF for accurate classification of segmentation points into valid and invalid achieved from heuristic segmentation. MLP and RBF are special types of Artificial Neural Network (ANN) composed of interconnected "neurons" which exchange messages among each other from inputs to outputs using single hidden layer. MLP and RBF are the standard learning algorithm from the same class of ANN use feed forward strategy from input to output neurons. However, the activation functions in each category operate differently. The both types have their own pros and cons; hence, from outcome of each model, weighted average is calculated to come out with better results and to cover limitations of each category for precise classification results [67,68]. Actually, both MLP and RBF needs training to solve a problem, in this phase networks establish a relation between inputs and outputs in conjunction with learning parameters. The networks map the relationship between the inputs and outputs, and then modify its internal functions to determine the best relationship. Accordingly, two networks (MLP and RBF) are trained and tested to evaluate best possible network structure, input neurons, output neurons and number of hidden layer(s). The structure of proposed ensemble model is exhibited in Figure 3.

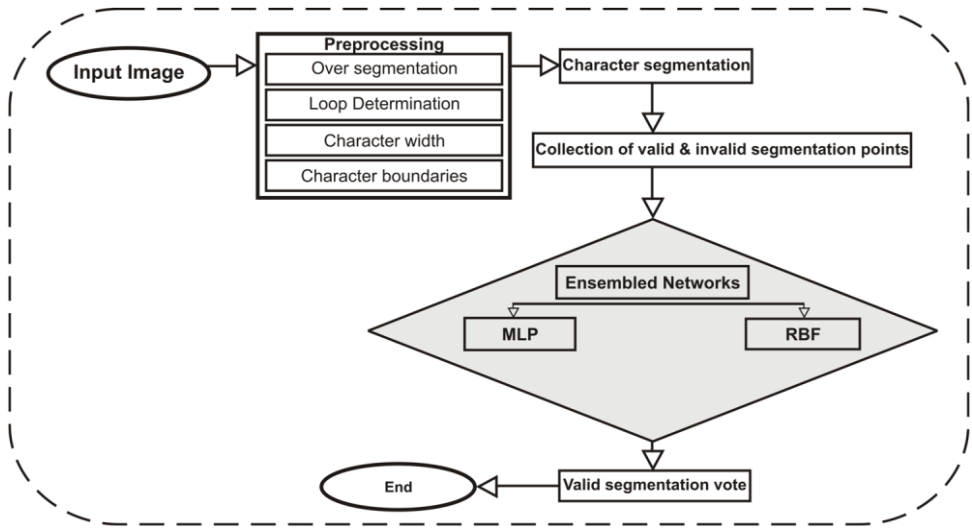

Fig 3. Proposed ensemble neural model

Initially, for training purpose all segmentation points are manually categorized into accurate and inaccurate segmentation points and are stored in one file that is used by individual networks and ensemble neural model. In fact, there are no standard criteria to calculate number of neurons in hidden layer, it is calculated based on trial and error based experiments by keeping MSE under observation. However, for the number of neurons in input and output layers depends on the strategy in the application. Initial weights in neural networks training are set at random that are altered based on decreased MSE. However, ensemble model calculates weighted average of these independent networks output to enhance the generalization ability. For training and testing ANN tool box available in MATLAB 7.0 is employed.

MLP, RBF and ensemble model are trained to come out with vote of confidence for valid and invalid character segmentation points to assists the segmentation process. Accordingly, several experiments are performed using different weights, number of epochs, momentum & learning rate. There are different procedures to train individual networks and to combine their output as well. However, training is stopped when both networks have identical configuration although their initial conditions were different. For the ensemble model, simple average of individual networks is calculated. Some neural networks training/testing results are exhibited in Appendix A

**2.2 Neural Validation**

In this step all segmentation points received from heuristic segmentation process are validated using trained neural network prepared in the previous step. These segmentation points are fed to the individual neural networks and ensemble model as well to classify all segmentation points into correct and incorrect based on confidence vote. It is observed that ensemble model exhibited better performance as compared to individual networks due to their limited generalization ability. Fig 4 exhibits the process of identification of invalid segmentation points that were voted negative using ensemble model.

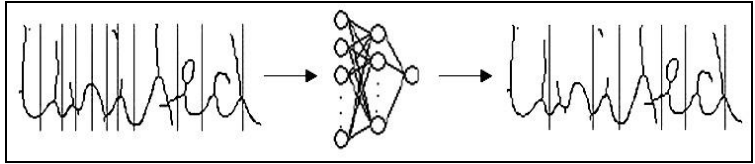

Fig 4. Invalid segmentation points are extracted using neural vote

**2.3 Non-linear Segmentation Path Detection**

Following, invalid segmentation point's identification using ensemble neural model and their removal, the next step is character segmentation without segmenting their discriminatory parts. Nonetheless, in the previous step, true character boundaries are identified. However, rather to perform simple vertical segmentation, nonlinear segmentation path is traced. Accordingly, first core-zone in each script word is detected [7] and valid segmentation points are traced up and down to the image boundaries such that it couldn't cut the characters body. Accordingly, segmentation path is traced according to the specified criteria. Few results are presented in Figure 5.

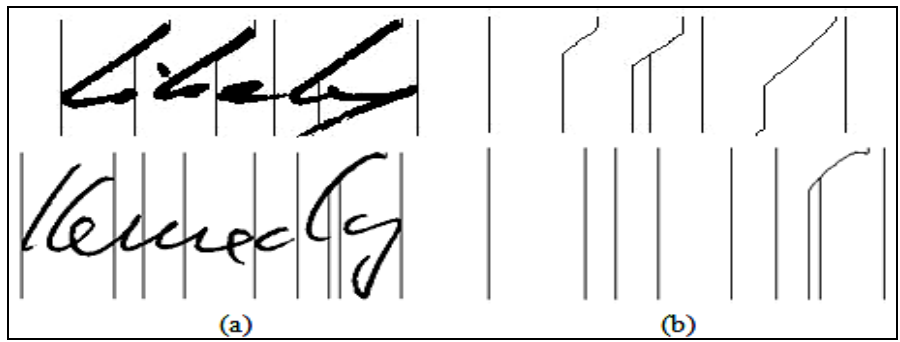

Fig 5: (a) Non-linear segmentation (b) segmentation path

3. **Analysis and Discussion**

**3.1 Handwriting database**

All experiments are conducted on training and test sets available in the CEDAR benchmark database [8]. Accordingly, some cursive script samples for characters segmentation, training and testing of the ensemble neural model are exhibited in fig 6.

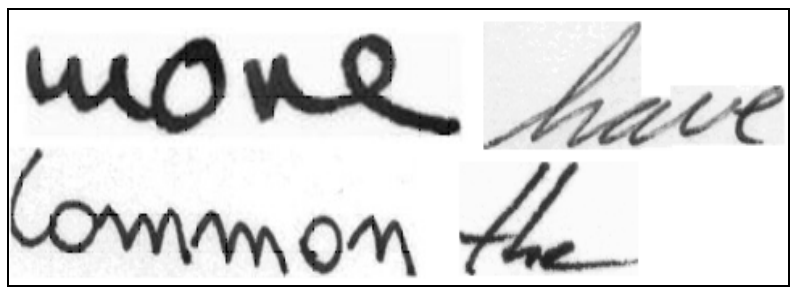

**Fig. 6**. Cursive Script word samples of CEDAR benchmark database

## 3.2 Experimental results

Various distinctive factual measurements accessible are utilized to assess the degree to which the qualities that have been anticipated are of good and sensible quality. Root mean square error (RMSE) is required frequently, to measure or assess the effectiveness of the neural ensemble model. The best neural structure is one that has less error and acceptable learning time. RMSE returns true values to be utilized as a standardized frame to compare the performance of the model on the expected and attained output. Finally, RMSE and R correlation coefficient between actual and target data, SI denotes scatter index.

$$RMSE = \sqrt{\frac{\sum_{i=1}^{N}(y_i - x_i)^2}{N}}$$

$$R = \frac{\sum_{i=1}^{N}(x_i - \overline{x})(y_i - \overline{y})}{\sqrt{\sum_{i=1}^{N}(x_i - \overline{x})^2 \sum_{i=1}^{N}(y_i - \overline{y})^2}}$$

$$SI = RMSE / \overline{x}$$

where $x_i$ represents attained values at the ith time step, $y_i$ is the attained values, N for counter increment, $\overline{x}$ and $\overline{y}$ represents actual and attained values.

Two sets of experiments are performed on datasets taken from CEDAR benchmark Datasets. The average recognition rates are calculated by adding all the recognition rates for each word using three step processes; (heuristic segmentation, an ensemble of neural networks assistance and non-linear segmentation strategy) and dividing the calculated sum by the number of words; Table 1 shows experimental results.

**Table 1. Character segmentation rates**

| Number of Words | Segmentation Points | Avg. Seg. Rate of Training set (%) | Avg. Seg. Testing Rate on Testing set (%) |
|---|---|---|---|
| 100 | 780 | 98.30 | 94.26 |
| 317 | 1829 | 100 | 97.98 |

### 3.3 Results Analysis and Discussion

Literature exhibits that researchers have employed various approaches for handwritten script segmentation into individual characters. Verma and Gader [69] achieved 91% segmentation accuracy with neural network feature based approach. The experiments are conducted on CEDAR samples without mentioning words number. In another study, conducted by Blumenstein and Verma [70] attained 78.85% segmentation accuracy with neuro-feature based approach on CEDAR samples. In the same line of action, Verma [71] noted 84.87% precision for segmentation of 300 words taken from CEDAR with neural assistance. In the same line, Cheng et al [72] utilized neural network for assistance and attained 95.27% segmentation rate from 317 CEDAR words. Cheng and Blumenstein [73] utilizing neuro-enhanced features based approach and claimed 84.19% character segmentation rate for 317

CEDAR words. However, all reported character segmentation techniques perform linear segmentation. Whilst, proposed approach produced 96.87% accuracy rate for CEDAR test set. In further work a much larger database shall be used for training the neural network. Segmentation results available in the literature since 2000 are presented in Table 2.

**Table 2**. Character segmentation accuracy rates on CEDAR

| Author | Segmentation method | Rate (%) | Segmentation Type | Comments |
| --- | --- | --- | --- | --- |
| Verma and Gader [69] | Feature based + Neural Network | 91 | Linear | Number of words for training / testing not mentioned |
| Blumenstein and Verma [70] | Feature based + Neural Network | 78.8 | Linear | Number of words for training / testing not mentioned |
| Verma [71] | Feature based + Neural Network | 84.87 | Linear | CEDAR test set |
| Cheng et al [72] | Feature based + Neural Network | 95.27 | Linear | CEDAR test set |
| Cheng and Blumenstein [73] | Enhanced feature based + Neural Network | 84.19 | Linear | CEDAR test set |
| Khan and Mohamad [40] | Geometric features + Neural Network | 91.21 | Linear | CEDAR test set |
| **Proposed approach** | Heuristic + Ensemble of Neural Networks | 97.53 | Non-Linear | CEDAR test set |

## 4. Conclusion

This paper has presented a heuristic character segmentation approach, a neural networks based ensemble model to assist character segmentation process and finally a non-linear segmentation strategy to enhance accuracy. However, in the first phase as a result of heuristic segmentation, few characters are over-segmented and heuristics are failed to come out with correct letter boundaries. In order to overcome this difficulty, a neural ensemble model is trained and integrated with the proposed approach. The ensemble neural model performance was better than individual neural networks and characters over-segmentation is significantly less hence, neural ensemble model was quite fast, that is an unusual observation. Non-linear segmentation strategy is also one of the main reasons behind the high character segmentation accuracy up to 97.53%; that is promising as compared to results reported in state of art. Finally, this paper has described solution to over-segmentation and to avoid linear segmentation successfully; while the problem of miss-segmentation for touched characters will be consider in future research.

**Appendix A** (ANN Training/ Test Results on CEDAR database)

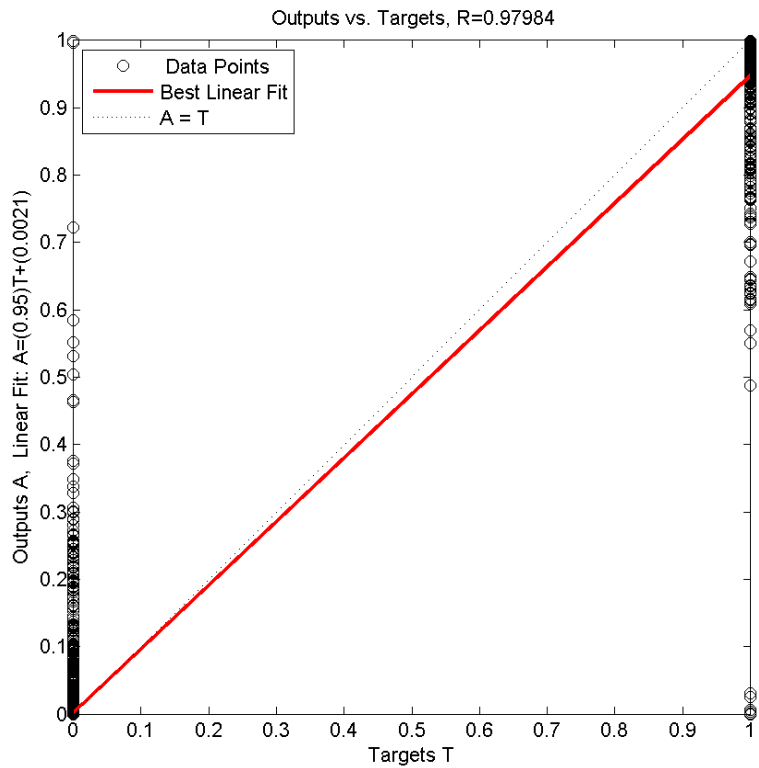

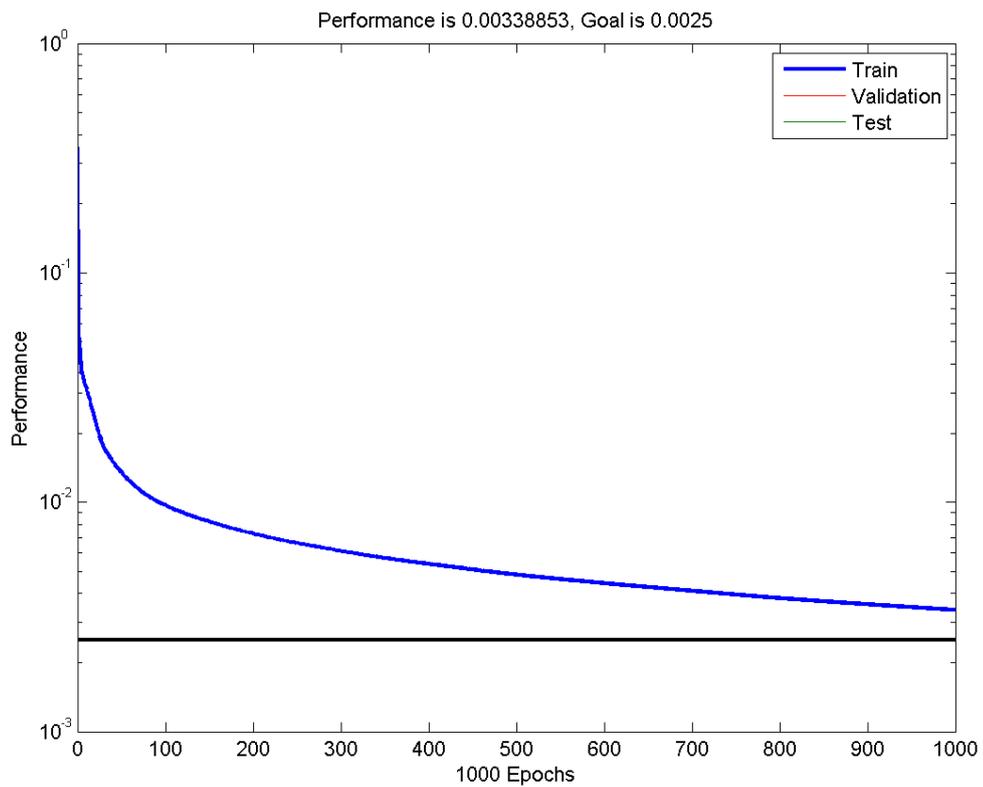

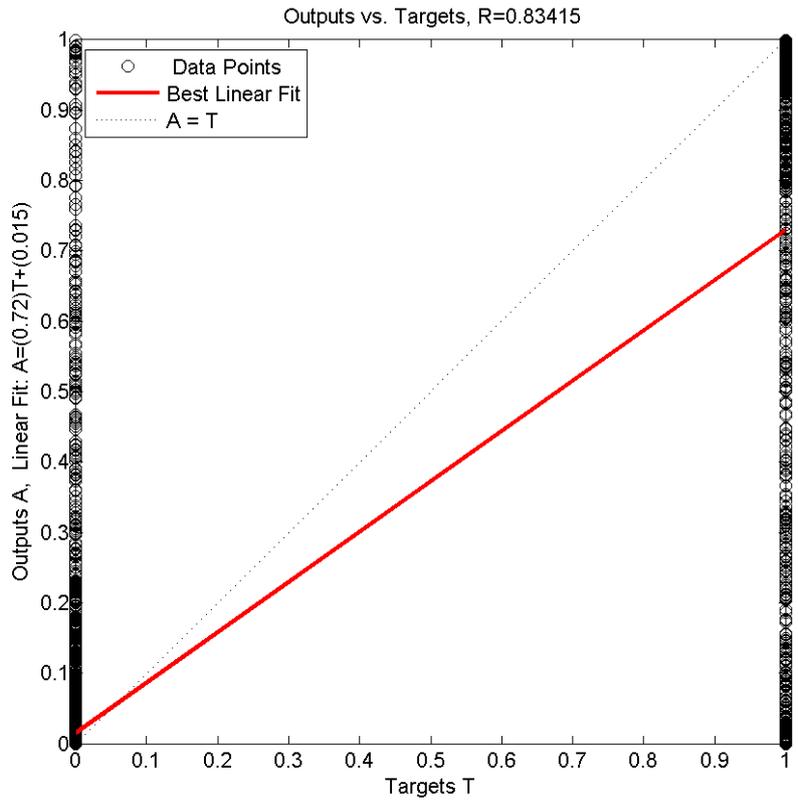
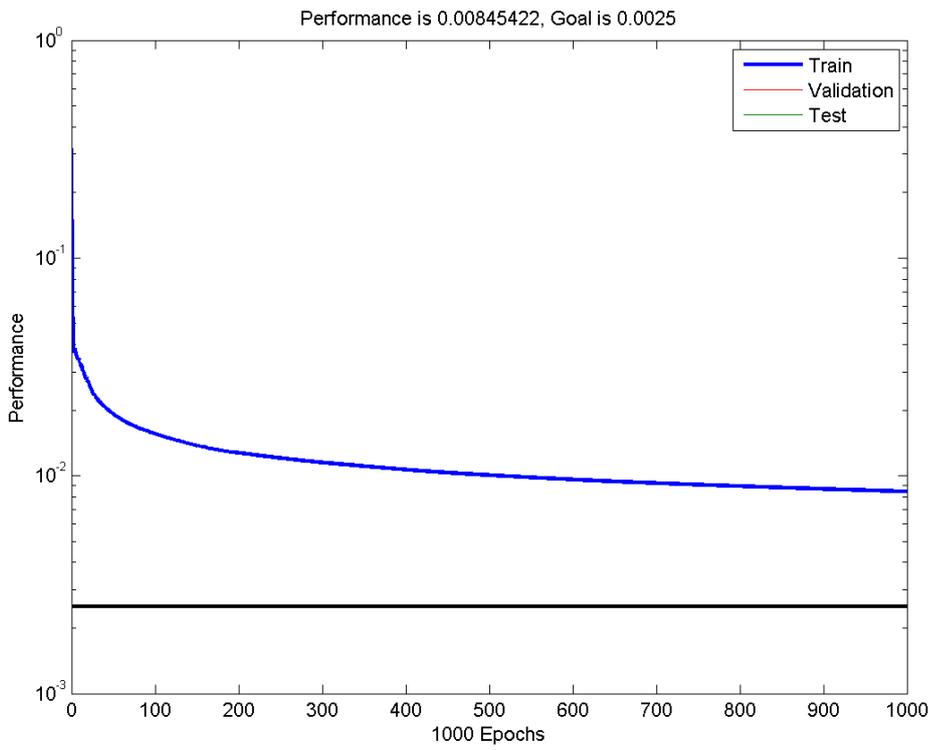

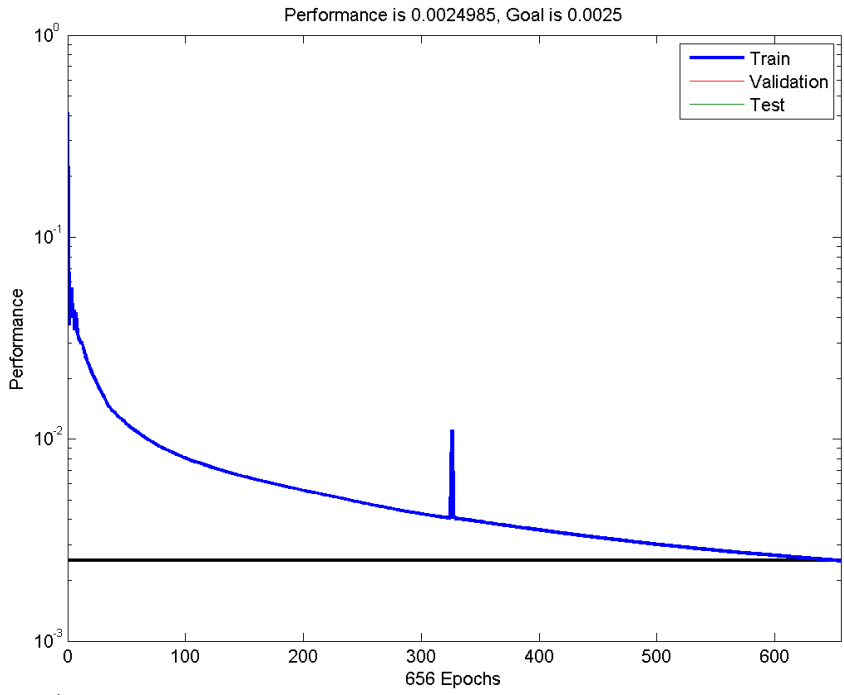

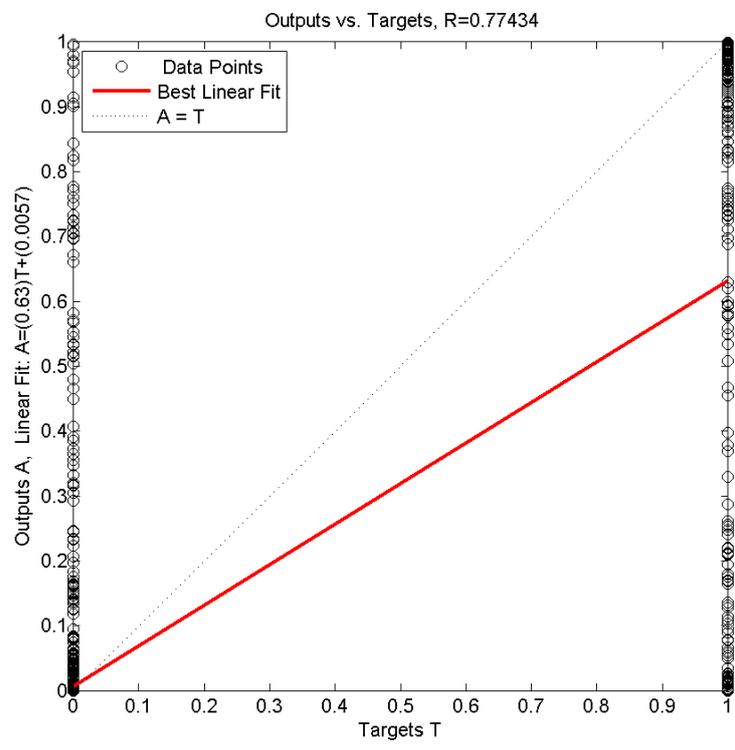